\documentclass{article}

\PassOptionsToPackage{numbers}{natbib}
\usepackage[final]{nips_2018}

\usepackage[utf8]{inputenc}
\usepackage[T1]{fontenc}
\usepackage{hyperref}
\usepackage{url}
\usepackage{booktabs}
\usepackage{amsfonts}
\usepackage{nicefrac}
\usepackage{microtype}

\usepackage{graphicx}
\graphicspath{{./figures/}}

\usepackage{subcaption}
\usepackage{wrapfig}

\usepackage{tikz,tikz-3dplot}
\usepackage{tikzscale}
\usepackage{pgfplots}
\usepgfplotslibrary{colormaps}
\pgfplotsset{compat=1.13}

  \usetikzlibrary{calc,fadings,decorations.pathmorphing,decorations.text,arrows,matrix,shapes,shapes.misc,patterns,plotmarks,positioning,intersections,3d,scopes,external}

  \newcommand{\includeTikZ}[2][]{
    \tikzsetnextfilename{#2}
    \includegraphics[#1]{./tikz_figures/#2.tikz}}

\usepackage{amsmath}
\usepackage{amssymb}
\usepackage{amsthm}
\usepackage{physics}
\usepackage{mathtools}
\usepackage{bm}
\usepackage{enumitem}
  
  \DeclareMathOperator*{\argmax}{arg\,max}
  \DeclareMathOperator{\st}{s.t.}

  \makeatletter
  \newcommand{\@giventhatstar}[2]{\left(#1\!\:\middle|\!\:#2\right)}
  \newcommand{\@giventhatnostar}[3][]{#1(#2\!\:#1|\!\:#3#1)}
  \newcommand{\giventhat}{\@ifstar\@giventhatstar\@giventhatnostar}
  \makeatother

  \def\x{\boldsymbol{x}}
  \def\xadv{\boldsymbol{x}_{\text{adv}}}
  \def\yadv{y_{\text{adv}}}
  
  \def\r{\boldsymbol{r}}
  \def\w{\boldsymbol{w}}

  \usepackage{algorithm}
  \usepackage{algorithmic}
  \usepackage{comment}
  \usepackage{cleveref}

\title{Improved Network Robustness\\ with Adversary Critic}

\author{
  Alexander Matyasko, Lap-Pui Chau\\
  School of Electrical and Electronic Engineering\\
  Nanyang Technological University, Singapore\\
  \texttt{aliaksan001@ntu.edu.sg, elpchau@ntu.edu.sg}
}

\begin{document}

\maketitle

\begin{abstract}
  Ideally, what confuses neural network should be confusing to humans. However,
  recent experiments have shown that small, imperceptible perturbations can
  change the network prediction. To address this gap in perception, we propose a
  novel approach for learning robust classifier. Our main idea is: adversarial
  examples for the robust classifier should be indistinguishable from the
  regular data of the adversarial target. We formulate a problem of learning
  robust classifier in the framework of Generative Adversarial Networks~(GAN),
  where the adversarial attack on classifier acts as a generator, and the critic
  network learns to distinguish between regular and adversarial images. The
  classifier cost is augmented with the objective that its adversarial examples
  should confuse the adversary critic. To improve the stability of the
  adversarial mapping, we introduce adversarial cycle-consistency constraint
  which ensures that the adversarial mapping of the adversarial examples is
  close to the original. In the experiments, we show the effectiveness of our
  defense. Our method surpasses in terms of robustness networks trained with
  adversarial training. Additionally, we verify in the experiments with human
  annotators on MTurk that adversarial examples are indeed visually confusing.
\end{abstract}

\section{Introduction}\label{sec:introduction}

Deep neural networks are powerful representation learning models which achieve
near-human performance in image~\cite{he2015deep} and
speech~\cite{hinton2012deep} recognition tasks. Yet, state-of-the-art networks
are sensitive to small input perturbations. \cite{szegedy2014intriguing} showed
that adding \textit{adversarial noise} to inputs produces images which are
visually similar to the images inputs but which the network misclassifies with
high confidence. In speech recognition, \cite{carlini2018audio} introduced an
adversarial attack, which can change any audio waveform, such that the corrupted
signal is over $99.9\%$ similar to the original but transcribes to any targeted
phrase. The existence of \textit{adversarial examples} puts into question
generalization ability of deep neural networks, reduces model interpretability,
and limits applications of deep learning in safety and security-critical
environments~\cite{sharif2016crime, papernot2016limitations}.

Adversarial training~\cite{goodfellow2014explaining,
  kurakin2016adverarialtraining, tramer2017ensemble} is the most popular
approach to improve network robustness. Adversarial examples are generated
online using the latest snapshot of the network parameters. The generated
adversarial examples are used to augment training dataset. Then, the classifier
is trained on the mixture of the original and the adversarial images. In this
way, adversarial training smoothens a decision boundary in the vicinity of the
training examples. Adversarial training~(AT) is an intuitive and effective
defense, but it has some limitations. AT is based on the assumption that
adversarial noise is label non-changing. If the perturbation is too large, the
adversarial noise may change the true underlying label of the input. Secondly,
adversarial training discards the dependency between the model parameters and
the adversarial noise. As a result, the neural network may fail to anticipate
changes in the adversary and overfit the adversary used during training.

\begin{figure}[!t]
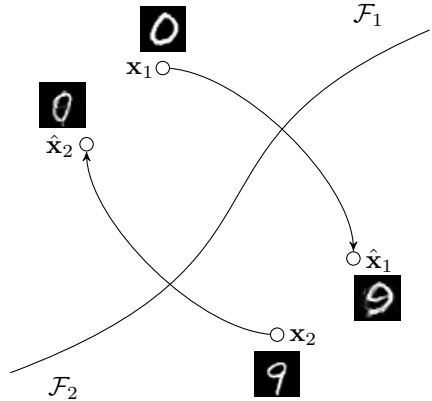

  \centering
  \includeTikZ[width=0.4\linewidth]{prior_diagram_b}%
  \vspace{-0.1cm}
  \caption{Adversarial examples should be indistinguishable from the regular
    data of the adversarial target. The images in the figure above are generated
    using~\citet{carlini2016towards} $l_2$-attack on the network trained with
    our defense, such that the confidence of the prediction on the adversarial
    images is $95\%$. The confidence on the original images $\x_1$ and $\x_2$ is
    $99\%$.~\label{fig:robust_prior}}
  \vspace{-0.15cm}
\end{figure}

Ideally, what confuses neural network should be confusing to humans. So the
changes introduced by the adversarial noise should be associated with removing
identifying characteristics of the original label and adding identifying
characteristics of the adversarial label. For example, images that are
adversarial to the classifier should be visually confusing to a human observer.
Current techniques~\cite{goodfellow2014explaining,
  kurakin2016adverarialtraining, tramer2017ensemble} improve robustness to input
perturbations from a selected uncertainty set. Yet, the model's adversarial
examples remain semantically meaningless. To address this gap in perception, we
propose a novel approach for learning robust classifier. Our core idea is that
adversarial examples for the robust classifier should be indistinguishable from
the regular data of the attack's target class~(see~\cref{fig:robust_prior}).

We formulate the problem of learning robust classifier in the framework of
Generative Adversarial Networks~(GAN)~\cite{goodfellow2014generative}. The
adversarial attack on the classifier acts as a generator, and the critic network
learns to distinguish between natural and adversarial images. We also introduce
a novel targeted adversarial attack which we use as the generator. The
classifier cost is augmented with the objective that its adversarial images
generated by the attack should confuse the adversary critic. The attack is
fully-differentiable and implicitly depends on the classifier parameters. We
train the classifier and the adversary critic jointly with backpropagation. To
improve the stability of the adversarial mapping, we introduce adversarial
cycle-consistency constraint which ensures that the adversarial mapping of the
adversarial examples is close to the original. Unlike adversarial training, our
method does not require adversarial noise to be label non-changing. To the
contrary, we require that the changes introduced by adversarial noise should
change the ``true'' label of the input to confuse the critic. In the
experiments, we demonstrate the effectiveness of the proposed approach. Our
method surpasses in terms of robustness networks trained with adversarial
training. Additionally, we verify in the experiments with human annotators that
adversarial examples are indeed visually confusing.

\section{Related work}\label{sec:related-work}

\textbf{Adversarial attacks} \hspace{2mm} \citet{szegedy2014intriguing} have
originally introduced a targeted adversarial attack which generates adversarial
noise by optimizing the likelihood of input for some adversarial target using a
box-constrained L-BFGS method. Fast Gradient Sign
method~(FGSM)~\cite{goodfellow2014explaining} is a one-step attack which uses a
first-order approximation of the likelihood loss. Basic Iterative Method~(BIM),
which is also known as Projected Gradient
Descent~(PGD),~\cite{kurakin2016adversarialexamples} iteratively applies the
first-order approximation and projects the perturbation after each step.
\cite{papernot2016limitations}~propose an iterative method which at each
iteration selects a single most salient pixel and perturbs it.
DeepFool~\cite{dezfooli2015deepfool} iteratively generates adversarial
perturbation by taking a step in the direction of the closest decision boundary.
The decision boundary is approximated with first-order Taylor series to avoid
complex non-convex optimization. Then, the geometric margin can be computed in
the closed-form. \citet{carlini2016towards} propose an optimization-based attack
on a modified loss function with implicit box-constraints.
\cite{papernot2016practical}~introduce a black-box adversarial attack based on
transferability of adversarial examples. Adversarial Transformation
Networks~(ATN)~\cite{baluja2017adversarial} trains a neural network to attack.

\textbf{Defenses against adversarial attacks} \hspace{2mm} Adversarial
training~(AT)~\cite{goodfellow2014explaining} augments training batch with
adversarial examples which are generated online using Fast Gradient Sign method.
Virtual Adversarial training~(VAT)~\cite{miyato2015distributional} minimizes
Kullback{-}Leibler divergence between the predictive distribution of clean
inputs and adversarial inputs. Notably, adversarial examples can be generated
without using label information and VAT was successfully applied in
semi-supervised settings. \cite{madry2017towards} applies iterative Projected
Gradient Descent~(PGD) attack to adversarial training. Stability
training~\cite{zheng2016improving} minimizes a task-specific distance between
the output on clean and the output on corrupted inputs. However, only a random
noise was used to distort the input. \cite{matyasko2017margin, elsayed2018large}
propose to maximize a geometric margin to improve classifier robustness.
Parseval networks~\cite{moustapha2017parseval} are trained with the
regularization constraint, so the weight matrices have a small spectral radius.
Most of the existing defenses are based on robust optimization and improve the
robustness to perturbations from a selected uncertainty set.

Detecting adversarial examples is an alternative way to mitigate the problem of
adversarial examples at test time. \cite{metzen2017detecting} propose to train a
detector network on the hidden layer's representation of the guarded model. If
the detector finds an adversarial input, an autonomous operation can be stopped
and human intervention can be requested. \cite{feinman2017detecting} adopt a
Bayesian interpretation of Dropout to extract confidence intervals during
testing. Then, the optimal threshold was selected to distinguish natural images
from adversarial. Nonetheless, \citet{carlini2017adversarial} have extensively
studied and demonstrated the limitations of the detection{-}based methods. Using
modified adversarial attacks, such defenses can be broken in both white-box and
black-box setups. In our work, the adversary critic is somewhat similar to the
adversary detector. But, unlike adversary-detection methods, we use information
from the adversary critic to improve the robustness of the guarded model during
training and do not use the adversary critic during testing.

\textbf{Generative Adversarial Networks}~\cite{goodfellow2014generative}
introduce a generative model where the learning problem is formulated as an
adversarial game between discriminator and generator. The discriminator is
trained to distinguish between real images and generated images. The generator
is trained to produce naturally looking images which confuse the discriminator.
A two-player minimax game is solved by alternatively optimizing two models.
Recently several defenses have been proposed which use GAN framework to improve
robustness of neural networks. Defense-GAN~\cite{samangouei2018defensegan} use
the generator at test time to project the corrupted input on the manifold of the
natural examples. \citet{lee2017generative} introduce Generative Adversarial
Trainer~(GAT) in which the generator is trained to attack the classifier. Like
Adversarial Training~\cite{goodfellow2014explaining}, GAT requires that
adversarial noise does not change the label. Compare with defenses based on
robust optimization, we do not put any prior constraint on the adversarial
attack. To the contrary, we require that adversarial noise for robust classifier
should change the ``true'' label of the input to confuse the critic. Our
formulation has three components (the classifier, the critic, and the attack)
and is also related to Triple-GAN~\cite{li2017triple}. But, in our work: 1) the
generator also fools the classifier; 2) we use the implicit dependency between
the model and the attack to improve the robustness of the classifier. Also, we
use a fixed algorithm to attack the classifier.

\section{Robust Optimization}\label{sec:perc-robustn}

We first recall a mathematical formulation for the robust multiclass
classification. Let $f(\x; \mathbf{W})$ be a $k$-class classifier, e.g. neural
network, where $\x \in \mathcal{R}^{N}$ is in the input space and $\mathbf{W}$
are the classifier parameters. The prediction rule is $\hat{k}(\x)=\argmax
f(\x)$. Robust optimization seeks a solution robust to the worst-case input
perturbations:
\begin{equation}
  \label{eq:robust_classification}
  \underset{\mathbf{W}}{\min} \, \underset{\r_{i} \in \mathcal{U}_{i}}{\max} \sum_{i=1}^{N} \mathcal{L}(f(\x_{i} + \r_{i}), y_{i})
\end{equation}
where $\mathcal{L}$ is a training loss, $\r_i$ is an arbitrary (even
adversarial) perturbation for the input $\x_{i}$, and $\mathcal{U}_i$ is an
uncertainty set, e.g. $l_{p}$-norm $\epsilon$-ball $\mathcal{U}_i=\{\r_{i}:
\norm{\r_{i}}_{p} \le \epsilon\}$. Prior information about the task can be used
to select a problem-specific uncertainty set $\mathcal{U}$.

Several regularization methods can be shown to be equivalent to the robust
optimization, e.g. $l_1$ lasso regression~\cite{xu2009robust} and $l_2$ support
vector machine~\cite{xu2009robustness}. Adversarial
training~\cite{goodfellow2014explaining} is a popular regularization method to
improve neural network robustness. AT assumes that adversarial noise is label
non-changing and trains neural network on the mixture of original and
adversarial images:
\begin{equation}
  \label{eq:adversarial_training}
  \underset{\mathbf{W}}{\min} \, \sum_{i=1}^{N} \mathcal{L}(f(\x_{i}), y_{i}) + \lambda \mathcal{L}(f(\x_{i} + \r_{i}), y_{i})
\end{equation}
where $\r_i$ is the adversarial perturbation generated using Fast Gradient Sign
method~(FGSM). \citet{shaham2015understanding} show that adversarial training is
a form of robust optimization with $l_{\infty}$-norm constraint.
\citet{madry2017towards} experimentally argue that Projected Gradient
Descent~(PGD) adversary is inner maximizer of~\cref{eq:robust_classification}
and, thus, PGD is the optimal first-order attack. Adversarial training with PGD
attack increases the robustness of the regularized models compare to the
original defense. Margin maximization~\cite{matyasko2017margin} is another
regularization method which generalizes SVM objective to deep neural networks,
and, like SVM, it is equivalent to the robust optimization with the margin loss.

\begin{wrapfigure}{R}{0.45\textwidth}
  \vspace{-1em}
  \centering
  \includegraphics[width=0.43\textwidth]{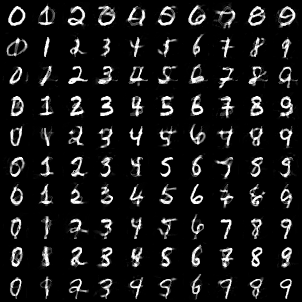}
  \vspace{-0.1em}
  \caption{Images on the diagonal are corrupted with the adversarial noise
    generated by CW~\cite{carlini2016towards} $l_2$-norm attack, so the
    prediction confidence on the adversarial images is at least $95\%$. The
    prediction confidence on the original images is
    $99\%$.~\label{fig:confusing_images}}
  \vspace{-1em}
\end{wrapfigure}

Selecting a good uncertainty set $\mathcal{U}$ for robust optimization is
crucial. Poorly chosen uncertainty set may result in an overly conservative
robust model. Most importantly, each perturbation $\r \in \mathcal{U}$ should
leave the ``true'' class of the original input $\x$ unchanged. To ensure that
the changes of the network prediction are indeed fooling examples,
\citet{goodfellow2014explaining} argue in favor of a max-norm perturbation
constraint for image classification problems. However, simple disturbance
models~(e.g.~$l_2$- and $l_{\infty}$-norm $\epsilon$-ball used in adversarial
training) are inadequate in practice because the distance to the decision
boundary for different examples may significantly vary. To adapt uncertainty set
to the problem at hand, several methods have been developed for constructing
data-dependent uncertainty sets using statistical hypothesis
tests~\cite{bertsimas2018datadriven}. In this work, we propose a novel approach
for learning a robust classifier which is orthogonal to prior robust
optimization methods.

Ideally, inputs that are adversarial to the classifier should be confusing to a
human observer. So the changes introduced by the adversarial noise should be
associated with the removing of identifying characteristics of the original
label and adding the identifying characteristics of the adversarial target. For
example, adversarial images in~\Cref{fig:confusing_images} are visually
confusing. The digit `$1$' (second row, eighth column) after adding the top
stroke was classified by the neural network as digit `$7$'. Likewise, the digit
`$7$' (eighth row, second column) after removing the top stroke was classified
by the network as digit `$1$'. Similarly for other images
in~\Cref{fig:confusing_images}, the model's ``mistakes'' can be predicted
visually. Such behavior of the classifier is expected and desired for the
problems in computer vision. Additionally, it improves the interpretability of
the model. In this work, we study image classification problems, but our
formulation can be extended to the classification tasks in other domains,
e.g.~audio or text.

Based on the above intuition, we develop a novel formulation for learning a
robust classifier. Classifier is robust if its adversarial examples are
indistinguishable from the regular data of the adversarial
target~(see~\cref{fig:robust_prior}). So, we formulate the following
mathematical problem:
\begin{equation}
  \label{eq:perceptual_robustness}
  \min \, \sum_{i=1}^{N} \mathcal{L}(f(\x_{i}), y_{i}) + \lambda \mathcal{D}\left[p_{\text{data}}\left(\x, y\right), p_{\text{adv}}\left(\x, y\right) \right]
\end{equation}
where $p_{\text{data}}\left(\x, y\right)$ and $p_{\text{adv}}\left(\x, y\right)$
is the distribution of the natural and the adversarial for $f$ examples and the
parameter $\lambda$ controls the trade-off between accuracy and robustness. Note
that the distribution $p_{\text{adv}}\left(\x, y\right)$ is constructed by
transforming natural samples $\left(\x, y\right) \sim p_{\text{data}}\left(\x,
  y\right)$ with $y \neq \yadv$, so that adversarial example $\xadv =
\mathcal{A}_{f}\left(\x; \yadv\right)$ is classified by $f$ as the attack's
target $\yadv$.

The first loss in~\cref{eq:perceptual_robustness}, e.g. NLL, fits the model
predictive distribution to the data distribution. The second term measures the
probabilistic distance between the distribution of the regular and adversarial
images and constrains the classifier, so its adversarial examples are
indistinguishable from the regular inputs. It is important to note that we
minimize a probabilistic distance between joint distributions because the
distance between marginal distributions $p_{\text{data}}(\x)$ and
$p_{\text{adv}}(\x)$ is trivially minimized when $\r \sim 0$. Compare with
adversarial training, the proposed formulation does not impose the assumption
that adversarial noise is label non-changing. To the contrary, we require that
adversarial noise for the robust classifier should be visually confusing and,
thus, it should change the underlying label of the input. Next, we will describe
the implementation details of the proposed defense.

\section{Robust Learning with Adversary Critic}\label{sec:learn-advers-exampl}

\begin{figure}[!b]
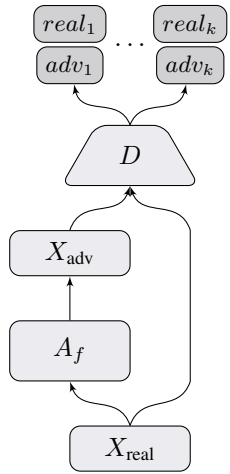

  \vspace{-1em}
  \begin{minipage}[!t]{0.42\linewidth}
    \begin{figure}[H]
      \centering
      \includeTikZ{adversary_critic}%
      \caption{Multiclass Adversary Critic.\label{fig:adversary_critic}}
    \end{figure}
  \end{minipage}%
  \begin{minipage}[!t]{0.58\linewidth}
    \begin{algorithm}[H]
      \begin{algorithmic}[1]
        \STATE \textbf{Input:} Image $\x$, target $y$, network $f$, confidence $C$.
        \STATE \textbf{Output:} Adversarial image $\hat{\x}$.
        \STATE $\hat{\x} \gets \x$
        \WHILE{$\ p_{y}(\hat{x}) < C$}
        \STATE $f \gets \log C - \log p_{y}(\hat{\x})$
        \STATE $\w \gets \nabla \log p_{y}(\hat{\x})$
        \STATE $\r \gets \frac{f}{\norm{\w}_{2}^2}\w$~\label{lst:line:clip_dist}
        \STATE $\hat{\x} \gets \hat{\x} + \r$
        \ENDWHILE
      \end{algorithmic}
      \caption{High-Confidence Attack $\mathcal{A}_f$ \label{alg:high_confidence_attack}}
    \end{algorithm}
  \end{minipage}
\end{figure}

As we have argued in the previous section, adversarial examples for the robust
classifier should be indistinguishable from the regular data of the adversarial
target. Minimizing the statistical distance between $p_{\text{data}} \left(\x,
  y\right)$ and $p_{\text{adv}} \left(\x, y\right)$
in~\cref{eq:perceptual_robustness} requires probability density estimation which
in itself is a difficult problem. Instead, we adopt the framework of Generative
Adversarial Networks~\cite{goodfellow2014generative}. We rely on a
discriminator, or \textit{an adversary critic}, to estimate a measure of
difference between two distributions. The discriminator given an input-label
pair $\left(\x, y\right)$ classifies it as either natural or adversarial. For
the $k$-class classifier $f$, we implement the adversary critic as a $k$-output
neural network~(see~\cref{fig:adversary_critic}). The objective for the $k$-th
output of the discriminator $D$ is to correctly distinguish between natural and
adversarial examples of the class $y_k$:
\begin{equation}
  \label{eq:critic_objective}
  \mathcal{L}(f^{*}, D_k) = \underset{D_k}{\min} \ \mathbb{E}_{\x\sim p_{\text{data}}\giventhat*{\x}{y_k}}  \left[ \log D_k\left(\x\right) \right] + \mathbb{E}_{y: y \neq y_k} \mathbb{E}_{\x\sim p_{\text{data}}\giventhat*{\x}{y}} \left[ \log \left( 1 - D_{k}\left(\mathcal{A}_{f^{*}}(\x; y_k) \right) \right) \right]
\end{equation}
where $\mathcal{A}_f(\x, y_k)$ is the targeted adversarial attack on the
classifier $f$ which transforms the input $\x$ to the adversarial target $y_k$.
An example of such attack is Projected Gradient
Descent~\cite{kurakin2016adversarialexamples} which iteratively takes a step in
the direction of the target $y_k$. Note that the second term
in~\cref{eq:critic_objective} is computed by transforming the regular inputs
$\left(\x, y\right) \sim p_{\text{data}}\left(\x, y\right)$ with the original
label $y$ different from the adversarial target $y_k$.

Our architecture for the discriminator in~\Cref{fig:adversary_critic} is
slightly different from the previous work on joint distribution
matching~\cite{li2017triple} where the label information was added as the input
to each layer of the discriminator. We use class label only in the final
classification layer of the discriminator. In the experiments, we observe that
with the proposed architecture: 1) the discriminator is more stable during
training; 2) the classifier $f$ converges faster and is more robust. We also
regularize the adversary critic with a gradient norm
penalty~\cite{gulrajani2017improved}. For the gradient norm penalty, we do not
interpolate between clean and adversarial images but simply compute the penalty
at the real and adversarial data separately. Interestingly, regularizing the
gradient of the binary classifier has the interpretation of maximizing the
geometric margin~\cite{matyasko2017margin}.

The objective for the classifier $f$ is to minimize the number of mistakes subject to
that its adversarial examples generated by the attack $\mathcal{A}_{f}$ fool the
adversary critic $D$:
\begin{equation}
  \label{eq:generator_objective}
  \mathcal{L}(f, D^{*}) = \underset{f}{\min} \ \mathbb{E}_{\x, y \sim p_{\text{data}}\left(x, y \right)}\mathcal{L}(f(\x), y) + \lambda \sum_{y_k} \mathbb{E}_{y: y \neq y_k} \mathbb{E}_{\x\sim p_{\text{data}}\giventhat*{\x}{y}} \left[ \log D_{k}^{*}\left(\mathcal{A}_{f}(\x; y_k) \right) \right]
\end{equation}
where $\mathcal{L}$ is a standard supervised loss such as negative
log-likelihood~(NLL) and the parameter $\lambda$ controls the trade-off between
test accuracy and classifier robustness. To improve stability of the adversarial
mapping during training, we introduce adversarial cycle-consistency constraint
which ensures that adversarial mapping $\mathcal{A}_f$ of the adversarial
examples should be close to the original:
\begin{equation}
  \label{eq:cycle_consistency}
  \mathcal{L}_{\text{cycle}}(y_s,y_t) = \mathbb{E}_{\x \sim p_{\text{data}}\giventhat*{\x}{y_s}}\left[ \norm{\mathcal{A}_{f}(\mathcal{A}_{f}(\x, y_{t}), y_s) - \x}_{2} \right] \; \forall y_s \neq y_t
\end{equation}
where $y_s$ is the original label of the input and $y_t$ is the adversarial
target. Adversarial cycle-consistency constraint is similar to cycle-consistency
constraint which was introduced for image-to-image
translation~\cite{zhu2017unpaired}. But, we introduce it to constraint the
adversarial mapping $\mathcal{A}_f$ and it improves the robustness of the
classifier $f$. Next, we discuss implementation of our targeted adversarial
attack $\mathcal{A}_f$.

Our defense requires that the adversarial attack $\mathcal{A}_f$ is
differentiable. Additionally, adversarial examples generated by the attack
$\mathcal{A}_f$ should be misclassified by the network $f$ with high confidence.
Adversarial examples which are close to the decision boundary are likely to
retain some identifying characteristics of the original class. An attack which
optimizes for the mistakes, e.g.~DeepFool~\cite{dezfooli2015deepfool},
guarantees the confidence of $\frac{1}{k}$ for $k$-way classifier. To generate
high-confidence adversarial examples, we propose a novel adversarial attack
which iteratively maximizes the confidence of the adversarial target. The
confidence of the target $k$ after adding perturbation $\r$ is $p_{k}(\x + \r)$.
The goal of the attack is to find the perturbation, so the adversarial input is
misclassified as $k$ with the confidence at least $C$:
\begin{equation*}
  \begin{split}
    \min \; & \norm{\r} \\
    \st \;  & p_{k}(\x + \r) \ge C
  \end{split}
\end{equation*}
We apply a first-order approximation to the constraint inequality:
\begin{equation*}
  \begin{split}
    \min \ & \norm{\r} \\
    \st \ & p_{k}(\x) + r \nabla_{\x} p_{k}(\x) \ge C
  \end{split}
\end{equation*}

Softmax in the final classification layer saturates quickly and shatters the
gradient. To avoid small gradients, we use log-likelihood instead. Finally, the
$l_{2}$-norm minimal perturbation can be computed using a method of Lagrange
multipliers as follows:
\begin{equation}
  \label{eq:minimal_perturbation}
  \r_k = \frac{\log C - \log p_{k}(\x)}{\norm{\nabla_{\x} \log p_{k}(\x)}_2}
\end{equation}
Because we use the approximation of the non-convex decision boundary, we
iteratively update perturbation $\r$ for $N_{\max}$ steps
using~\cref{eq:minimal_perturbation} until the adversarial input $\xadv$ is
misclassified as the target $k$ with the confidence $C$. Our attack can be
equivalently written as $\xadv = \x + \prod_{i=1}^{N_{\max}} I(p(\x +
\sum_{j=1}^{i} \r_j) \leq C) \r_i$ where $I$ is an indicator function. The
discrete stopping condition introduces a non-differentiable path in the
computational graph. We replace the gradient of the indicator function $I$ with
sigmoid-adjusted straight-through estimator during
backpropagation~\cite{bengio2013estimating}. This is a biased estimator but it
has low variance and performs well in the experiments.

The proposed attack is similar to Basic Iterative Method
(BIM)~\cite{kurakin2016adversarialexamples}. BIM takes a fixed $\epsilon$-norm
step in the direction of the attack target while our method uses an adaptive
step $\gamma = \frac{\abs{\log C - \log p_y(\hat{\x})}}{\norm{\nabla_x \log
    p_y(\hat{\x})}}$. The difference is important for our defense:
\vspace{-0.5em}
\begin{enumerate}
\item BIM introduces an additional parameter $\epsilon$. If $\epsilon$ is too
  large, then the attack will not be accurate. If $\epsilon$ is too small, then
  the attack will require many iterations to converge.
\item Both attacks are differentiable. However, for BIM attack during
  backpropagation, all the gradients $\frac{\partial \r_i}{\partial \w}$ have an
  equal weight $\epsilon$. For our attack, the gradients will be weighted
  adaptively depending on the distance $\gamma$ to the attack's target. The step
  $\gamma$ for our attack is also fully-differentiable.
\end{enumerate}
\vspace{-0.5em}

Full listing of our attack is shown in~\cref{alg:high_confidence_attack}. Next,
we discuss how we select the adversarial target $y_t$ and the attack's target
confidence $C$ during training.

The classifier $f$ approximately characterizes a conditional distribution $p
\giventhat*{y}{\x}$. If the classifier $f^*$ is optimal and robust, its
adversarial examples generated by the attack $\mathcal{A}_f$ should fool the
adversary critic $D$. Therefore, the attack $\mathcal{A}_f$ to fool the critic
$D$ should generate adversarial examples with the confidence $C$ equal to the
confidence of the classifier $f$ on the regular examples. During training, we
maintain a running mean of the confidence score for each class on the regular
data. The attack target $y_t$ for the input $\x$ with the label $y_s$ can be
sampled from the masked uniform distribution. Alternatively, the class with the
closest decision boundary~\cite{dezfooli2015deepfool} can be selected. The
latter formulation resulted in more robust classifier $f$ and we used it in all
our experiments. This is similar to support vector machine formulation which
maximizes the minimum margin.

Finally, we train the classifier $f$ and the adversary critic $D$ jointly using
stochastic gradient descent by alternating minimization of
\Cref{eq:critic_objective,eq:generator_objective}. Our formulation has three
components (the classifier $f$, the critic $D$, and the attack $\mathcal{A}_f$)
and it is similar to Triple-GAN~\cite{li2017triple} but the generator in our
formulation also fools the classifier.

\section{Experiments}\label{sec:experiments}

Adversarial training~\cite{goodfellow2014explaining} discards the dependency
between the model parameters and the adversarial noise. In this work, it is
necessary to retain the implicit dependency between the classifier $f$ and the
adversarial noise, so we can backpropagate through the adversarial attack
$\mathcal{A}_f$. For these reasons, all experiments were conducted using
Tensorflow~\cite{ttdt2016tensorflow} which supports symbolic differentiation and
computation on GPU. Backpropagation through our attack requires second-order
gradients $\frac{\partial^2 f(\x;\w)}{\partial x \partial w}$ which increases
computational complexity of our defense. At the same time, this allows the model
to anticipate the changes in the adversary and, as we show, significantly
improves the model robustness both numerically and perceptually. Codes for the
project are available
\href{https://github.com/aam-at/adversary_critic}{on-line}.

We perform experiments on MNIST dataset. While MNIST is a simple classification
task, it remains unsolved in the context of robust learning. We evaluate
robustness of the models against $l_2$ attacks. Minimal adversarial perturbation
$\r$ is estimated using DeepFool~\cite{dezfooli2015deepfool},
\citet{carlini2016towards}, and the proposed attack. To improve the accuracy of
DeepFool and our attack during testing, we clip the $l_2$-norm of perturbation
at each iteration to $0.1$. Note that our attack with the fixed step is
equivalent to Basic Iterative Method~\cite{kurakin2016adversarialexamples}. We
set the maximum number of iterations for DeepFool and our attack to $500$. The
target confidence $C$ for our attack is set to the prediction confidence on the
original input $\x$. DeepFool and our attack do not handle domain constraints
explicitly, so we project the perturbation after each update. For
\citet{carlini2016towards}, we use implementation provided by the authors with
default settings for the attack but we reduce the number of optimization
iterations from $10000$ to $1000$. As suggested in~\cite{dezfooli2015deepfool},
we measure the robustness of the model as follows:
\begin{equation}\label{eq:average_robustness}
  \rho_{\text{adv}}(\mathcal{A}_{f}) = \frac{1}{\lvert\mathcal{D}\rvert} \sum_{\mathbf{x} \in \mathcal{D}} \frac{\norm{\mathbf{r}(\mathbf{x})}_2}{\norm{\mathbf{x}}_2}
\end{equation}
where $\mathcal{A}_f$ is the attack on the classifier $f$ and $\mathcal{D}$ is
the test set.

We compare our defense with reference (no defense), Adversarial
Training~\cite{goodfellow2014explaining,
  kurakin2016adverarialtraining}~($\epsilon = 0.1$), Virtual Adversarial
Training~(VAT)~\cite{miyato2015distributional}~($\epsilon = 2.0$), and
$l_{2}$-norm Margin Maximization~\cite{matyasko2017margin}~($\lambda = 0.1$)
defense. We study the robustness of two networks with rectified activation: 1) a
fully-connected neural network with three hidden layers of size $1200$ units
each; 2) Lenet-5 convolutional neural network. We train both networks using Adam
optimizer~\cite{kingma2014adam} with batch size $100$ for $100$ epochs. Next, we
will describe the training details for our defense.

Our critic has two layers with $1200$ units each and leaky rectified activation.
We also add Gaussian noise to the input of each layer. We train both the
classifier and the critic using Adam~\cite{kingma2014adam} with the momentum
$\beta_1=0.5$. The starting learning rate is set to $5 \cdot 10^{-4}$ and
$10^{-3}$ for the classifier and the discriminator respectively. We train our
defense for $100$ epochs and the learning rate is halved every $40$ epochs. We
set $\lambda = 0.5$ for fully-connected network and $\lambda = 0.1$ for Lenet-5
network which we selected using validation dataset. Both networks are trained
with $\lambda_{\text{rec}}=10^{-2}$ for the adversarial cycle-consistency loss
and $\lambda_{\text{grad}}=10.0$ for the gradient norm penalty. The number of
iterations for our attack $\mathcal{A}_f$ is set to $5$. The attack confidence
$C$ is set to the running mean class confidence of the classifier on natural
images. We pretrain the classifier $f$ for $1$ epoch without any regularization
to get an initial estimate of the class confidence scores.

\begin{table}[!b]
  \vspace{-0.7em}
  \renewcommand{\arraystretch}{1.1}
  \begin{subtable}{0.5\linewidth}
    \centering
    \begin{tabular}[t]{l|l|l|l|l}
      \hline
      Defense   & $\%$ & \cite{dezfooli2015deepfool} & \cite{carlini2016towards} & Our \\
      \hline
      Reference                       & $1.46$ & $0.131$          & $0.124$          & $0.173$\\
      \cite{goodfellow2014explaining} & $0.90$ & $0.228$          & $0.210$          & $0.299$\\
      \cite{miyato2015distributional} & $0.84$ & $0.244$          & $0.215$          & $0.355$\\
      \cite{matyasko2017margin}       & $0.84$ & $0.262$          & $0.230$          & $0.453$\\
      Our                             & $1.18$ & $\mathbf{0.290}$ & $\mathbf{0.272}$ & $\mathbf{0.575}$ \\
      \hline
    \end{tabular}
    \vspace{-0.3em}
    \caption{\label{tab:mlp_mnist_results}}
  \end{subtable}%
  \begin{subtable}{0.5\linewidth}
    \centering
    \begin{tabular}[t]{l|l|l|l|l}
      \hline
      Defense   & $\%$ & \cite{dezfooli2015deepfool} & \cite{carlini2016towards} & Our \\
      \hline
      Reference                       & $0.64$ & $0.157$          & $0.148$          & $0.207$ \\
      \cite{goodfellow2014explaining} & $0.55$ & $0.215$          & $0.191$          & $0.286$ \\
      \cite{miyato2015distributional} & $0.60$ & $0.225$          & $0.195$          & $0.330$ \\
      \cite{matyasko2017margin}       & $0.54$ & $0.248$          & $0.225$          & $0.470$ \\
      Our                             & $0.93$ & $\mathbf{0.288}$ & $\mathbf{0.278}$ & $\mathbf{0.590}$ \\
      \hline
    \end{tabular}
    \vspace{-0.3em}
    \caption{\label{tab:lenet5_mnist_results}}
  \end{subtable}
  \caption{Results on MNIST dataset for fully-connected network
    in~\cref{tab:mlp_mnist_results} and for Lenet-5 convolutional network
    in~\cref{tab:lenet5_mnist_results}. Column 1: test error on original images.
    Column 3-5: robustness $\rho$ under DeepFool~\cite{dezfooli2015deepfool},
    \citet{carlini2016towards}, and the proposed
    attack.~\label{tab:mnist_results}}
  \vspace{-1em}
\end{table}

\begin{figure}[t!]
  \centering
  \begin{subfigure}{0.32\linewidth}
    \includegraphics[width=\linewidth]{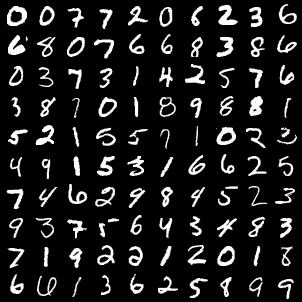}\caption{\label{fig:orig_adversarial_images}}
  \end{subfigure}\hfill
  \begin{subfigure}{0.32\linewidth}
    \includegraphics[width=\linewidth]{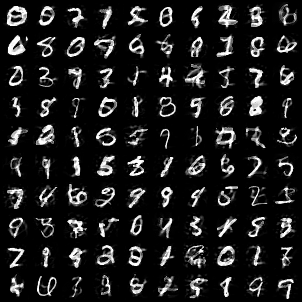}\caption{\label{fig:adversarial_images_carlini}}
  \end{subfigure}\hfill
  \begin{subfigure}{0.32\linewidth}
    \includegraphics[width=\linewidth]{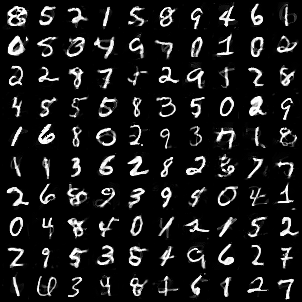}\caption{\label{fig:adversarial_images_carlini_prob9}}
  \end{subfigure}
  \caption{\Cref{fig:orig_adversarial_images} shows a random subset of test
    images~(average confidence $97\%$). \Cref{fig:adversarial_images_carlini}
    shows adversarial examples at the class decision boundary~(average
    confidence $34\%$). \Cref{fig:adversarial_images_carlini_prob9} shows
    high-confidence adversarial images (average confidence
    $98\%$).~\label{fig:adversarial_images}}
  \vspace{-0.5em}
\end{figure}

Our results for $10$ independent runs are summarized
in~\Cref{tab:mnist_results}, where the second column shows the test error on the
clean images, and the subsequent columns compare the robustness $\rho$ to
DeepFool~\cite{dezfooli2015deepfool}, \citet{carlini2016towards}, and our
attacks. Our defense significantly increases the robustness of the model to
adversarial examples. Some adversarial images for the neural network trained
with our defense are shown in~\Cref{fig:adversarial_images}. Adversarial
examples are generated using \citet{carlini2016towards} attack with default
parameters. As we can observe, adversarial examples at the decision boundary
in~\Cref{fig:adversarial_images_carlini} are visually confusing. At the same
time, high-confidence adversarial examples
in~\Cref{fig:adversarial_images_carlini_prob9} closely resemble natural images
of the adversarial target. We propose to investigate and compare various
defenses based on how many of its adversarial ``mistakes'' are actual mistakes.

We conduct an experiment with human annotators on MTurk. We asked the workers to
label adversarial examples. Adversarial examples were generated from the test
set using the proposed attack. The attack's target was set to class closest to
the decision boundary and the target confidence was set to the model's
confidence on the original examples. We split $10000$ test images into $400$
assignments. Each assignment was completed by one unique annotator. We report
the results for four defenses in~\Cref{tab:mturk_mnist_results}. For the model
trained without any defense, adversarial noise does not change the label of the
input. When the model is trained with our defense, the high-confidence
adversarial noise actually changes the label of the input.

\begin{table}[!t]
  \renewcommand{\arraystretch}{1.1}
  \begin{subtable}{0.5\linewidth}
    \centering
    \begin{tabular}[t]{l|l|l}
      \hline
      Defense   & $\%$ Change & $\%$ No change \\
      \hline
      Reference                       & $0.57$  & $98.74$ \\
      \cite{goodfellow2014explaining} & $19.02$ & $77.21$ \\
      \cite{miyato2015distributional} & $35.08$ & $59.68$ \\
      \cite{matyasko2017margin}       & $60.47$ & $34.52$ \\
      Our                             & $87.99$ & $9.86$ \\
      \hline
    \end{tabular}
    \caption{\label{tab:mlp_mturk_results}}
  \end{subtable}
  \begin{subtable}{0.5\linewidth}
    \centering
    \begin{tabular}[t]{l|l|l}
      \hline
      Defense   & $\%$ Change & $\%$ No change \\
      \hline
      Reference                       & $2.54$  & $96.53$ \\
      \cite{goodfellow2014explaining} & $19.1$  & $75.94$ \\
      \cite{miyato2015distributional} & $26.8$  & $67.73$ \\
      \cite{matyasko2017margin}       & $81.77$ & $13.15$ \\
      Our                             & $92.29$ & $6.51$ \\
      \hline
    \end{tabular}
    \caption{\label{tab:lenet5_mturk_results}}
  \end{subtable}
  \vspace{-0.5em}
  \caption{Results of Amazon Mechanical Turk experiment for fully-connected
    network in~\cref{tab:mlp_mturk_results} and for Lenet-5 convolutional
    network in~\cref{fig:adversarial_images_carlini_prob9}. Column 2: shows
    percent of adversarial images which human annotator label with its
    adversarial target, so adversarial noise changed the ``true'' label of the
    input. Column 3: shows percent of the adversarial images which human
    annotator label with its original label, so adversarial noise did not change
    the underlying label of the input.~\label{tab:mturk_mnist_results}}
  \vspace{-1em}
\end{table}

\section{Conclusion}\label{sec:conclusion}

In this paper, we introduce a novel approach for learning a robust classifier.
Our defense is based on the intuition that adversarial examples for the robust
classifier should be indistinguishable from the regular data of the adversarial
target. We formulate a problem of learning robust classifier in the framework of
Generative Adversarial Networks. Unlike prior work based on robust optimization,
our method does not put any prior constraints on adversarial noise. Our method
surpasses in terms of robustness networks trained with adversarial training. In
experiments with human annotators, we also show that adversarial examples for our
defense are indeed visually confusing. In the future work, we plan to scale our
defense to more complex datasets and apply it to the classification tasks in
other domains, such as audio or text.

\subsubsection*{Acknowledgments}
This work was carried out at the Rapid-Rich Object Search (ROSE) Lab at Nanyang
Technological University (NTU), Singapore. The ROSE Lab is supported by the
National Research Foundation, Singapore, and the Infocomm Media Development
Authority, Singapore. We thank NVIDIA Corporation for the donation of the
GeForce Titan X and GeForce Titan X (Pascal) used in this research. We also
thank all the anonymous reviewers for their valuable comments and suggestions.


\begin{thebibliography}{36}
\providecommand{\natexlab}[1]{#1}
\providecommand{\url}[1]{\texttt{#1}}
\expandafter\ifx\csname urlstyle\endcsname\relax
  \providecommand{\doi}[1]{doi: #1}\else
  \providecommand{\doi}{doi: \begingroup \urlstyle{rm}\Url}\fi

\bibitem[He et~al.(2016)He, Zhang, Ren, and Sun]{he2015deep}
K.~He, X.~Zhang, S.~Ren, and J.~Sun.
\newblock Deep residual learning for image recognition.
\newblock In \emph{CVPR}, 2016.

\bibitem[Hinton et~al.(2012)Hinton, Deng, Yu, Dahl, r.~Mohamed, Jaitly, Senior,
  Vanhoucke, Nguyen, Sainath, and Kingsbury]{hinton2012deep}
G.~Hinton, L.~Deng, D.~Yu, G.~E. Dahl, A.~r.~Mohamed, N.~Jaitly, A.~Senior,
  V.~Vanhoucke, P.~Nguyen, T.~N. Sainath, and B.~Kingsbury.
\newblock Deep neural networks for acoustic modeling in speech recognition: The
  shared views of four research groups.
\newblock In \emph{IEEE Signal Processing Magazine}, 2012.

\bibitem[{Szegedy} et~al.(2013){Szegedy}, {Zaremba}, {Sutskever}, {Bruna},
  {Erhan}, {Goodfellow}, and {Fergus}]{szegedy2014intriguing}
C.~{Szegedy}, W.~{Zaremba}, I.~{Sutskever}, J.~{Bruna}, D.~{Erhan},
  I.~{Goodfellow}, and R.~{Fergus}.
\newblock {Intriguing properties of neural networks}.
\newblock In \emph{ICLR}, 2013.

\bibitem[{Carlini} and {Wagner}(2018)]{carlini2018audio}
N.~{Carlini} and D.~{Wagner}.
\newblock {Audio Adversarial Examples: Targeted Attacks on Speech-to-Text}.
\newblock \emph{arXiv preprint arXiv:1801.01944}, 2018.

\bibitem[Sharif et~al.(2016)Sharif, Bhagavatula, Bauer, and
  Reiter]{sharif2016crime}
Mahmood Sharif, Sruti Bhagavatula, Lujo Bauer, and Michael~K. Reiter.
\newblock Accessorize to a crime: Real and stealthy attacks on state-of-the-art
  face recognition.
\newblock In \emph{Proceedings of the 2016 ACM SIGSAC Conference on Computer
  and Communications Security}, 2016.

\bibitem[Papernot et~al.(2016)Papernot, McDaniel, Jha, Fredrikson, Celik, and
  Swami]{papernot2016limitations}
Nicolas Papernot, Patrick~D. McDaniel, Somesh Jha, Matt Fredrikson, Z.~Berkay
  Celik, and Ananthram Swami.
\newblock The limitations of deep learning in adversarial settings.
\newblock In \emph{{IEEE} European Symposium on Security and Privacy}, 2016.

\bibitem[Goodfellow et~al.(2015)Goodfellow, Shlens, and
  Szegedy]{goodfellow2014explaining}
Ian~J Goodfellow, Jonathon Shlens, and Christian Szegedy.
\newblock Explaining and harnessing adversarial examples.
\newblock In \emph{ICLR}, 2015.

\bibitem[{Kurakin} et~al.(2017){Kurakin}, {Goodfellow}, and
  {Bengio}]{kurakin2016adverarialtraining}
A.~{Kurakin}, I.~{Goodfellow}, and S.~{Bengio}.
\newblock {Adversarial Machine Learning at Scale}.
\newblock In \emph{ICLR}, 2017.

\bibitem[Tramèr et~al.(2018)Tramèr, Kurakin, Papernot, Goodfellow, Boneh, and
  McDaniel]{tramer2017ensemble}
Florian Tramèr, Alexey Kurakin, Nicolas Papernot, Ian Goodfellow, Dan Boneh,
  and Patrick McDaniel.
\newblock Ensemble adversarial training: Attacks and defenses.
\newblock In \emph{ICLR}, 2018.

\bibitem[Carlini and Wagner(2017{\natexlab{a}})]{carlini2016towards}
Nicholas Carlini and David~A. Wagner.
\newblock Towards evaluating the robustness of neural networks.
\newblock In \emph{{IEEE} Symposium on Security and Privacy}, 2017.

\bibitem[Goodfellow et~al.(2014)Goodfellow, Pouget{-}Abadie, Mirza, Xu,
  Warde{-}Farley, Ozair, Courville, and Bengio]{goodfellow2014generative}
Ian~J. Goodfellow, Jean Pouget{-}Abadie, Mehdi Mirza, Bing Xu, David
  Warde{-}Farley, Sherjil Ozair, Aaron~C. Courville, and Yoshua Bengio.
\newblock Generative adversarial nets.
\newblock In \emph{NIPS}, 2014.

\bibitem[{Kurakin} et~al.(2016){Kurakin}, {Goodfellow}, and
  {Bengio}]{kurakin2016adversarialexamples}
A.~{Kurakin}, I.~{Goodfellow}, and S.~{Bengio}.
\newblock {Adversarial examples in the physical world}.
\newblock \emph{arXiv preprint arXiv:1607.02533}, 2016.

\bibitem[Moosavi-Dezfooli et~al.(2016)Moosavi-Dezfooli, Fawzi, and
  Frossard]{dezfooli2015deepfool}
S.~M. Moosavi-Dezfooli, A.~Fawzi, and P.~Frossard.
\newblock Deepfool: A simple and accurate method to fool deep neural networks.
\newblock In \emph{CVPR}, 2016.

\bibitem[Papernot et~al.(2017)Papernot, McDaniel, Goodfellow, Jha, Celik, and
  Swami]{papernot2016practical}
Nicolas Papernot, Patrick~D. McDaniel, Ian~J. Goodfellow, Somesh Jha, Z.~Berkay
  Celik, and Ananthram Swami.
\newblock Practical black-box attacks against machine learning.
\newblock In \emph{Proceedings of the 2017 {ACM} on Asia Conference on Computer
  and Communications Security}, 2017.

\bibitem[Baluja and Fischer(2018)]{baluja2017adversarial}
Shumeet Baluja and Ian Fischer.
\newblock Learning to attack: Adversarial transformation networks.
\newblock In \emph{AAAI}, 2018.

\bibitem[{Miyato} et~al.(2015){Miyato}, {Maeda}, {Koyama}, {Nakae}, and
  {Ishii}]{miyato2015distributional}
T.~{Miyato}, S.-i. {Maeda}, M.~{Koyama}, K.~{Nakae}, and S.~{Ishii}.
\newblock {Distributional Smoothing with Virtual Adversarial Training}.
\newblock In \emph{ICLR}, 2015.

\bibitem[{Madry} et~al.(2017){Madry}, {Makelov}, {Schmidt}, {Tsipras}, and
  {Vladu}]{madry2017towards}
A.~{Madry}, A.~{Makelov}, L.~{Schmidt}, D.~{Tsipras}, and A.~{Vladu}.
\newblock {Towards Deep Learning Models Resistant to Adversarial Attacks}.
\newblock In \emph{ICLR}, 2018.

\bibitem[Zheng et~al.(2016)Zheng, Song, Leung, and
  Goodfellow]{zheng2016improving}
S.~Zheng, Y.~Song, T.~Leung, and I.~Goodfellow.
\newblock Improving the robustness of deep neural networks via stability
  training.
\newblock In \emph{CVPR}, 2016.

\bibitem[Matyasko and Chau(2017)]{matyasko2017margin}
Alexander Matyasko and Lap{-}Pui Chau.
\newblock Margin maximization for robust classification using deep learning.
\newblock In \emph{IJCNN}, 2017.

\bibitem[Elsayed et~al.(2018)Elsayed, Krishnan, Mobahi, Regan, and
  Bengio]{elsayed2018large}
Gamaleldin~Fathy Elsayed, Dilip Krishnan, Hossein Mobahi, Kevin Regan, and Samy
  Bengio.
\newblock Large margin deep networks for classification.
\newblock In \emph{NIPS}, 2018.

\bibitem[Ciss{\'{e}} et~al.(2017)Ciss{\'{e}}, Bojanowski, Grave, Dauphin, and
  Usunier]{moustapha2017parseval}
Moustapha Ciss{\'{e}}, Piotr Bojanowski, Edouard Grave, Yann Dauphin, and
  Nicolas Usunier.
\newblock Parseval networks: Improving robustness to adversarial examples.
\newblock In \emph{ICML}, 2017.

\bibitem[{Hendrik Metzen} et~al.(2017){Hendrik Metzen}, {Genewein}, {Fischer},
  and {Bischoff}]{metzen2017detecting}
J.~{Hendrik Metzen}, T.~{Genewein}, V.~{Fischer}, and B.~{Bischoff}.
\newblock {On Detecting Adversarial Perturbations}.
\newblock In \emph{ICLR}, 2017.

\bibitem[{Feinman} et~al.(2017){Feinman}, {Curtin}, {Shintre}, and
  {Gardner}]{feinman2017detecting}
R.~{Feinman}, R.~R. {Curtin}, S.~{Shintre}, and A.~B. {Gardner}.
\newblock {Detecting Adversarial Samples from Artifacts}.
\newblock \emph{arXiv preprint arXiv:1703.00410}, 2017.

\bibitem[Carlini and Wagner(2017{\natexlab{b}})]{carlini2017adversarial}
Nicholas Carlini and David~A. Wagner.
\newblock Adversarial examples are not easily detected: Bypassing ten detection
  methods.
\newblock In \emph{Proceedings of the 10th {ACM} Workshop on Artificial
  Intelligence and Security}, 2017.

\bibitem[Samangouei et~al.(2018)Samangouei, Kabkab, and
  Chellappa]{samangouei2018defensegan}
Pouya Samangouei, Maya Kabkab, and Rama Chellappa.
\newblock Defense-{GAN}: Protecting classifiers against adversarial attacks
  using generative models.
\newblock In \emph{ICLR}, 2018.

\bibitem[{Lee} et~al.(2017){Lee}, {Han}, and {Lee}]{lee2017generative}
H.~{Lee}, S.~{Han}, and J.~{Lee}.
\newblock {Generative Adversarial Trainer: Defense to Adversarial Perturbations
  with GAN}.
\newblock \emph{arXiv preprint arXiv:1705.03387}, 2017.

\bibitem[LI et~al.(2017)LI, Xu, Zhu, and Zhang]{li2017triple}
Chongxuan LI, Taufik Xu, Jun Zhu, and Bo~Zhang.
\newblock Triple generative adversarial nets.
\newblock In \emph{NIPS}, 2017.

\bibitem[Xu et~al.(2009{\natexlab{a}})Xu, Caramanis, and Mannor]{xu2009robust}
Huan Xu, Constantine Caramanis, and Shie Mannor.
\newblock Robust regression and lasso.
\newblock In \emph{NIPS}, 2009{\natexlab{a}}.

\bibitem[Xu et~al.(2009{\natexlab{b}})Xu, Caramanis, and
  Mannor]{xu2009robustness}
Huan Xu, Constantine Caramanis, and Shie Mannor.
\newblock Robustness and regularization of support vector machines.
\newblock In \emph{Journal of Machine Learning Research}, 2009{\natexlab{b}}.

\bibitem[{Shaham} et~al.(2015){Shaham}, {Yamada}, and
  {Negahban}]{shaham2015understanding}
U.~{Shaham}, Y.~{Yamada}, and S.~{Negahban}.
\newblock {Understanding Adversarial Training: Increasing Local Stability of
  Neural Nets through Robust Optimization}.
\newblock \emph{arXiv preprint arXiv:1511.05432}, 2015.


\bibitem[Bertsimas et~al.(2018)Bertsimas, Gupta, and
  Kallus]{bertsimas2018datadriven}
Dimitris Bertsimas, Vishal Gupta, and Nathan Kallus.
\newblock Data-driven robust optimization.
\newblock In \emph{Mathematical Programming}, 2018.

\bibitem[Gulrajani et~al.(2017)Gulrajani, Ahmed, Arjovsky, Dumoulin, and
  Courville]{gulrajani2017improved}
Ishaan Gulrajani, Faruk Ahmed, Martin Arjovsky, Vincent Dumoulin, and Aaron~C
  Courville.
\newblock Improved training of wasserstein gans.
\newblock In \emph{NIPS}, 2017.

\bibitem[Zhu et~al.(2017)Zhu, Park, Isola, and Efros]{zhu2017unpaired}
J.~Zhu, T.~Park, P.~Isola, and A.~A. Efros.
\newblock Unpaired image-to-image translation using cycle-consistent
  adversarial networks.
\newblock In \emph{ICCV}, 2017.

\bibitem[Bengio et~al.(2013)Bengio, L{\'{e}}onard, and
  Courville]{bengio2013estimating}
Yoshua Bengio, Nicholas L{\'{e}}onard, and Aaron~C. Courville.
\newblock Estimating or propagating gradients through stochastic neurons for
  conditional computation.
\newblock \emph{arXiv preprint arXiv:1308.3432}, 2015.

\bibitem[{Abadi} et~al.(2016){Abadi}, {Agarwal}, {Barham}, {Brevdo}, {Chen},
  {Citro}, {Corrado}, {Davis}, {Dean}, {Devin}, {Ghemawat}, {Goodfellow},
  {Harp}, {Irving}, {Isard}, {Jia}, {Jozefowicz}, {Kaiser}, {Kudlur},
  {Levenberg}, {Mane}, {Monga}, {Moore}, {Murray}, {Olah}, {Schuster},
  {Shlens}, {Steiner}, {Sutskever}, {Talwar}, {Tucker}, {Vanhoucke},
  {Vasudevan}, {Viegas}, {Vinyals}, {Warden}, {Wattenberg}, {Wicke}, {Yu}, and
  {Zheng}]{ttdt2016tensorflow}
The Tensorflow Development Team.
\newblock {TensorFlow: Large-Scale Machine Learning on Heterogeneous
  Distributed Systems}.
\newblock \emph{arXiv preprint arXiv:1603.04467}, 2015.

\bibitem[{Kingma} and {Ba}(2014)]{kingma2014adam}
D.~P. {Kingma} and J.~{Ba}.
\newblock {Adam: A Method for Stochastic Optimization}.
\newblock \emph{arXiv preprint arXiv:1412.6980}, 2015.

\end{thebibliography}
\end{document}